\documentclass[lettersize,journal]{IEEEtran}
\usepackage{amsmath,amsfonts}
\usepackage{algorithmic}
\usepackage{algorithm}
\usepackage{array}
\usepackage[caption=false,font=normalsize,labelfont=sf,textfont=sf]{subfig}
\usepackage{textcomp}
\usepackage{stfloats}
\usepackage{url}
\usepackage{verbatim}
\usepackage{graphicx}
\usepackage{cite}
\usepackage{soul, color, xcolor}
\usepackage{colortbl}
\usepackage{booktabs}

\usepackage{amsmath}
\usepackage{algorithm}
\usepackage{algorithmic}
\usepackage{amsfonts}
\usepackage{bbding}

\soulregister\cite7
\soulregister\eqref7
\begin{document}

\title{DGNR: Density-Guided Neural Point Rendering of Large Driving Scenes}

\author{
    Zhuopeng Li, Chenming Wu, Liangjun Zhang, Jianke Zhu~\IEEEmembership{Senior Member, IEEE}

\thanks{Zhuopeng Li is with the College of Software and Jianke Zhu is with the College of Computer Science, both at Zhejiang University, 38 Zheda Road, Hangzhou, China. 
Email: \texttt{\{lizhuopeng, jkzhu\}@zju.edu.cn};}
\thanks{Chenming Wu and Liangjun Zhang are with Robotics and Autonomous Driving Lab (RAL), Baidu Research. 
Email: \texttt{\{wuchenming, liangjunzhang\}@baidu.com}.}
\thanks{Jianke Zhu is the Corresponding Author.}
}

\maketitle

\begin{abstract}
Despite the recent success of Neural Radiance Field (NeRF), it is still challenging to render large-scale driving scenes with long trajectories, particularly when the rendering quality and efficiency are in high demand. Existing methods for such scenes usually involve with spatial warping, geometric supervision from zero-shot normal or depth estimation, or scene division strategies, where the synthesized views are often blurry or fail to meet the requirement of efficient rendering. To address the above challenges, this paper presents a novel framework that learns a density space from the scenes to guide the construction of a point-based renderer, dubbed as \textbf{DGNR (Density-Guided Neural Rendering)}. In DGNR, geometric priors are no longer needed, which can be intrinsically learned from the density space through volumetric rendering. Specifically, we make use of a differentiable renderer to synthesize images from the neural density features obtained from the learned density space. A density-based fusion module and geometric regularization are proposed to optimize the density space. By conducting experiments on a widely used autonomous driving dataset, we have validated the effectiveness of DGNR in synthesizing photorealistic driving scenes and achieving real-time capable rendering. Code will be released.

\end{abstract}


\section{Introduction}

\begin{figure}[t]
\centering
\includegraphics[width=\linewidth]{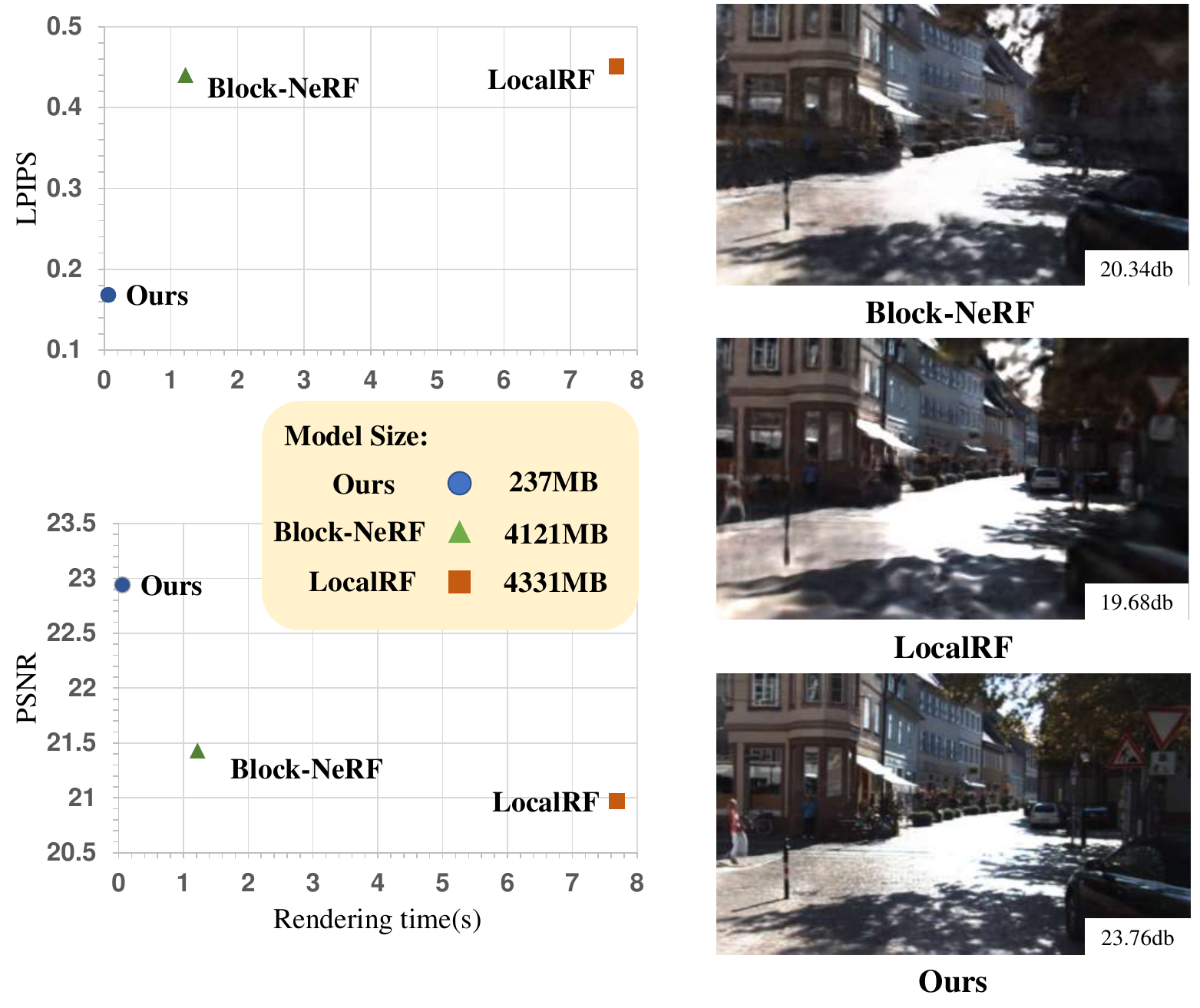}  
\caption{Novel views synthesized by NeRF-based methods tend to exhibit blurriness and suffer from noticeable artifacts. In contrast, DGNR focuses on synthesizing novel views with enhanced texture details, resulting in significantly improved visual quality. Moreover, our method has compact 3D representation and efficient rendering speed, which is over 20 times faster (16.67 FPS) than Block-NeRF~\cite{tancik2022block} (0.82 FPS) and LocalRF~\cite{meuleman2023progressively} (0.13 FPS).}
\label{fig:label1}
\end{figure}

\IEEEPARstart{A}{utonomous} driving (AD) has emerged as a significant technological breakthrough, leveraging deep learning networks to achieve remarkable advancements. It holds great promise and has the potential to revolutionize the future of transportation. However, ensuring the safety of autonomous driving systems has become a primary focus of extensive development efforts. Simulation has gained prominence as a reliable, secure, and efficient alternative for training and evaluating autonomous driving (AD) software and algorithms~\cite{li2019aads,amini2020learning, amini2022vista,nishida2022dynamic,chai2022deep,cai2023occupancy,lin2023immesh,placed2023survey,horvath2022object,lefevre2015learning}. It offers a robust platform that enables comprehensive testing and assessment of AD systems in a controlled virtual environment.

In the realm of neural rendering, notable progress has been achieved, with a prominent example being the remarkable strides made by Neural Radiance Fields (NeRF)~\cite{mildenhall2021nerf} in recent period. NeRF~\cite{sun2023nerf,tseng2022cla,yen2022nerf,zhu2023latitude,maggio2023loc,byravan2023nerf2real,adamkiewicz2022vision,ran2023neurar} has showcased remarkable abilities in achieving photorealistic reconstruction and synthesizing novel viewpoints. Nevertheless, challenges persist when it comes to capturing driving scenes, primarily due to the heightened complexity involved and the substantial computational resources necessary for achieving precise representation.
Due to the limited coverage by cameras, driving scenes present a unique challenge for neural representations. The distribution of camera locations tends to be biased towards driving modes rather than fully covering specific regions of interest. Consequently, only a small portion of the scene is observed. To address this challenge, research efforts such as Urban-NeRF~\cite{rematas2022urban} and S-NeRF~\cite{xie2023s} incorporate LiDAR data to supervise the depth estimation of NeRF, facilitating the learning of scene geometry~\cite{wang2023digging}. Additionally, Neural Point Light Field~\cite{ost2022neural} leverages LiDAR data to encode the radiance field and exploits the sparse geometry in point clouds. However, these methods heavily rely on the availability of LiDAR data or require strong geometric priors.
In this paper, we aim to address the challenges in outdoor driving scenes by proposing a novel approach that eliminates the reliance on LiDAR data or geometric priors, offering a more efficient and effective solution for neural rendering in these complex scenarios.
In order to tackle the issue of depending on geometric priors while preserving real-time rendering performance, we identify several key challenges: 1) generating urban views for extensive trajectories solely based on RGB image supervision without prior depth or additional supervision; 2) developing a compact 3D representation of the scene that enables real-time rendering of driving scenes; 3) mitigating artifacts like aliasing and floating objects, and synthesizing driving scenes with high-quality photorealistic textures.

To enable the real-time synthesis of driving scenes, our approach DGNR (Density-Guided Neural Rendering) involves with learning the density space through volume rendering and considering it as a geometric scene representation. By making use of differentiable rendering techniques, the density space of the 3D scene is expressed as neural density features in 2D space. Furthermore, a neural renderer generates images using the neural density features obtained from the initial 3D density space. Notably, the density space is learnable and can dynamically adapt its geometry through iterative feedback from rendered images.
To enhance the accuracy of the density space and facilitate the synthesis of photorealistic images, we divide the scene into blocks. This partitioning enables the density space to capture fine details of the scene. Additionally, we introduce a fusion module based on the density space, which merges multiple density spaces to ensure smoothness at the boundaries. Finally, we propose a depth smoothing regularization method to repair holes in the urban scene representation. Consequently, our approach enables real-time capable rendering of high-quality images with long trajectories in driving scenes.
The main contributions of this paper can be summarized as follows.

\begin{itemize}
    \item We propose a density-guided scene representation to compactly encode scenes using explicit representation, enabling the construction of large-scale driving scenes. Unlike existing approaches, our method does not rely on geometric priors, which achieves real-time performance.
    \item We further optimize the rendering quality based on the proposed scene representation by a new density-guided differentiable rendering method. This enhances the capabilities of synthesizing photorealistic images.
    \item Our extensive experiments demonstrate the effectiveness of our proposed method in representing long trajectories of driving scenes compared to other methods, achieving state-of-the-art in rendering large-scale driving scenes. 
\end{itemize}

\section{Related Work}

\subsection{Driving View Simulation}
In recent years, the adoption of autonomous driving simulation has gained significant popularity due to its valuable applications, such as verifying planning and control systems, generating training and testing data, and reducing time requirements for these tasks. Currently, two main types of simulators are utilized: model-based and data-driven.
Model-based simulators, such as PyBullet~\cite{coumans2016pybullet}, MuJoCo~\cite{todorov2012mujoco}, and CARLA~\cite{dosovitskiy2017carla}, rely on computer graphics techniques to simulate vehicles and environments. However, the manual process involved in creating these models and defining vehicle movements can be labor-intensive and time-consuming. Additionally, the resulting images may not always achieve the desired level of realism, which can lead to compromised performance when deploying perception systems.
On the other hand, data-driven simulators like AADS~\cite{li2019aads} and VISTA~\cite{amini2022vista, amini2020learning} address these challenges by utilizing real-world datasets to generate fully annotated and photorealistic simulations suitable for training and testing AD systems. These simulators employ driving view synthesis algorithms primarily based on conventional projection-based methods. More recently, there have been works that superficially simulate driving views~\cite{wu2023mapnerf, wu2023mars, yang2023unisim} built upon the NeRF technique, excelling in synthesizing photorealistic images and outperforming conventional view synthesis algorithms in AD simulation.

 \begin{figure*}[t]
\centering
  \includegraphics[width=\linewidth]{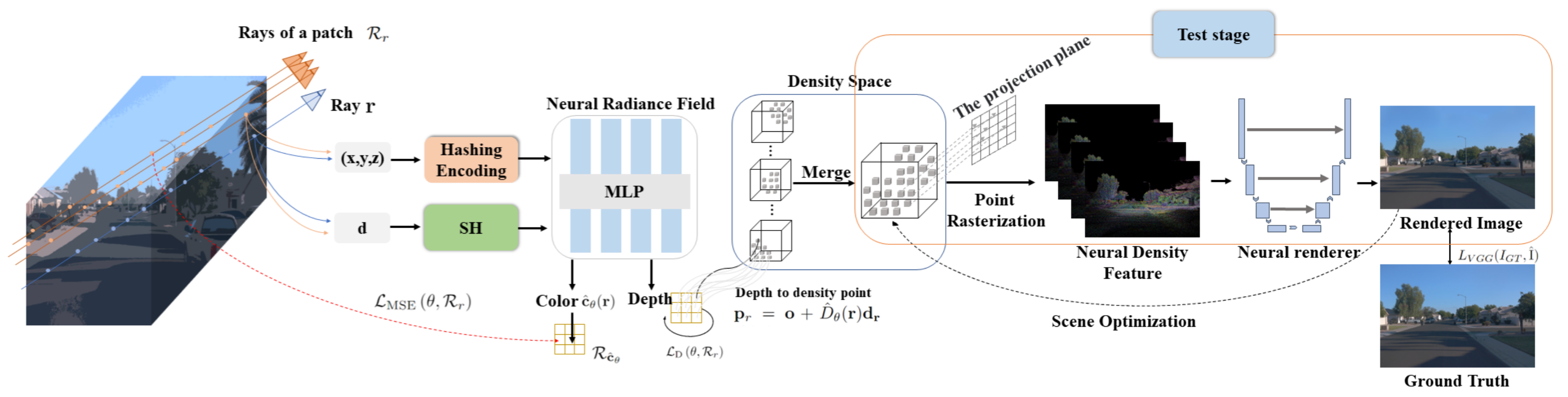}  
  \caption{Overview of DGNR. Firstly, Ray positions and directions are encoded using hashing and spherical harmonics encoding(SH). RGB images are employed to supervise the predicted colors, and depth smoothing regularization is imposed to optimize density space. Next, the learnable multiple density spaces are fused into a single-density space by a density fusion module, which serves as the geometric representation of the scene. Through rasterization, the 3D density space is projected into neural density features. Finally, a differentiable neural renderer is used to synthesize photorealistic driving scenes.}
  \label{fig:label2}
\end{figure*} 

\subsection{Novel View Synthesis of Driving Scenes}

NeRF offers effective scene representation but faces challenges in rendering large-scale driving scenes. NeRF++\cite{zhang2020nerf++} models foreground and background separately, Urban-NeRF\cite{rematas2022urban} uses a spherical environment map for the sky, Mip-NeRF 360~\cite{barron2022mip} and MeRF~\cite{reiser2023merf} map scene coordinate to a bounded volume, and F2-NeRF~\cite{wang2023f2} reduces capacity waste with a space-warping scheme. However, these methods struggle with long trajectory scenes, resulting in blurry and low-detail renderings.
The objective of novel view synthesis technology is to generate new views of objects or scenes from multiple images with different poses, enabling users to observe them from various perspectives. While NeRF-based methods have shown promising results in view synthesis for object-centric scenes rather than driving scenes.
To address these challenges, several approaches have been proposed in recent years. Using reprojection confidence, S-NeRF~\cite{xie2023s} introduces LiDAR points to learn robust geometry and address depth outliers. Neural Point Light Fields~\cite{ost2022neural} implicitly represent scenes through a light field residing on a sparse point cloud, efficiently encoding features with a single radiance evaluation per ray. Additionally, point-based neural rendering methods~\cite{li2023read, ruckert2022adop, kopanas2021point,kerbl20233d} have gained popularity by leveraging point cloud data as input to learn urban scene representations. These methods capture local geometric shapes and appearances by learning neural descriptors. These approaches take into account LiDAR, point cloud data, or external supervision.
Our work aims to propose a novel approach that takes advantage of existing methods to overcome the challenges specific to the neural rendering of driving scenes.

\subsection{Scalable Novel View Synthesis}

Recently, there have been notable advancements in reconstructing radiance fields for large-scale scenes by decomposing them into blocks and training distinct Neural Radiance Fields for each block, as demonstrated in the literature~\cite{tancik2022block, turki2022mega, turki2023suds}. These approaches effectively model large-scale scenes with long trajectories. Additionally, the LocalRF method~\cite{meuleman2023progressively} introduces a progressive scheme to allocate local radiance fields dynamically. 
However, the sampled points of each ray are greatly increased in large-scale driving scenes. Especially for the block division-based method, it is necessary to calculate the density and color of the scene from multiple blocks. The rendering speed of these methods is limited by the expression based on NeRFs. Our proposed DGNR approach, leveraging the Density-Guided Scene Representation, deviates from the original radiance field during rendering and employs differentiable rendering techniques for real-time rendering of large-scale scenes. This novel approach not only overcomes the limitations of previous methods but also enables the efficient synthesis of high-quality driving scenes in real-time.

\section{Method}
Our proposed method aims to address the challenges of rendering novel views of large-scale driving scenes. One of our key contributions is the density-guided scene representation, which leverages the capability of differentiable volumetric rendering to intrinsically construct the scene geometry. This representation is further refined by the proposed learnable density-based fusion module, where the details in the scene representation are significantly improved. The underlying geometric entity in our scene representation is the neural density point so that the whole scene could be efficiently rendered with a point-based rasterization approach in a differentiable manner. DGNR addresses the challenges associated with long trajectory unbounded driving scenes, achieving real-time rendering capabilities while capturing realistic texture details.

\subsection{Neural Radiance Field}

We first give preliminaries of neural radiance fields and define some notations of DGNR.
NeRF~\cite{mildenhall2021nerf} parameterizes a 3D location $\mathbf{x} \in \mathbb{R}^3$ and direction $\mathbf{d} \in \mathbb{S}^3$,  in which a radiance field is a continuous function $f$ mapping location and direction to a volume density $\sigma\in[0,\infty)$ and color value $\mathbf{c}\in[0,1]^{3}$. The volume density is predicted as a function of 3D position. NeRF and its variants use multi-layer perceptrons (MLPs), Neural Graphic Primitives (NGPs), triplanes, or voxels to represent a 3D space optimized by providing input images with poses.
\begin{equation}
\begin{array}{r}
f_\theta: \mathbb{R}^3 \times \mathbb{S}^2 \rightarrow[0,1]^3 \times [0,\infty).
\end{array}
\end{equation}

The sigma  $\sigma_\theta$ and color $\mathbf{c}_\theta$ indicate the density and color prediction of the radiance field using MLPs $f_\theta$, parameterized by $\theta$:
\begin{equation}
   \mathbf{c}_\theta, \sigma_\theta= f_\theta(\gamma(\mathbf{x}),\gamma(\mathbf{d})),
\end{equation}
where $\gamma$ represents a predefined spherical harmonics encoding (SH)~\cite{chen2022tensorf} or positional encoding.
Given a neural radiance field, a pixel is rendered by casting a ray $\mathbf{r}(u)=\mathbf{o}+u \mathbf{d}$ from the camera center $\mathbf{o}$ through the pixel along direction $\mathbf{d}$. Considering the predefined near and far planes for rendering $u_n$ and $u_f$, $Q(u)$ denotes the accumulated transmittance along the ray. The predicted color value $\hat{\mathbf{c}}_\theta$ is computed by alpha composition as below
\begin{equation}
\hat{\mathbf{c}}_\theta(\mathbf{r})=\int_{u_n}^{u_f} Q(u) \sigma_\theta(\mathbf{r}(u)) \mathbf{c}_\theta(\mathbf{r}(u), \mathbf{d}) d u ,
\end{equation}

\begin{equation}
 Q(u)=\exp \left(-\int_{u_n}^u \sigma_\theta(\mathbf{r}(s)) d s\right).
\end{equation}

A neural radiance field is optimized over a set of input images and their camera poses by minimizing the mean squared error as follows
\begin{equation}
\mathcal{L}_{\mathrm{MSE}}\left(\theta, \mathcal{R}_r\right)=\sum_{\mathbf{r} \in \mathcal{R}_r}\left\|\hat{\mathbf{c}}_\theta(\mathbf{r})-\mathbf{c}(\mathbf{r})\right\|^2,
\label{eq:label4}
\end{equation}
where $\mathcal{R}_r$ indicates a set of input rays, and $\mathbf{c}$ is ground truth color.
%
%
\subsection{Density-Guided Scene Representation}
In the context of long trajectory scenes that contain multiple objects of interest, the representational capacity of volumetric fields becomes constrained, leading to a decline in image quality during the rendering process. Therefore, we derive the scene representation from the density space of the radiance field. This allows us to use rasterization technology to represent the density space of 3D scenes as neural density features in 2D space, alleviating the pressure on the radiance field to express fine 3D scenes. The neural density features are then rendered into realistic driving scenes through a neural renderer. Next, we will explain how to obtain the density-guided scene representation and geometric optimization in this section.
Given the ray set $\mathcal{R}$, we can infer the predicted density point by 
\begin{equation}
\mathbf{p}_{r}=\mathbf{o}+\hat{\mathbf{D}}_\theta(\mathbf{r})\mathbf{d}_{\mathbf{r}}, 
\label{eq:label6}.
\end{equation}
where $\mathbf{p}_{r}$ is a 3D coordinate in world space. This helps us construct a density-guided scene representation $\mathbf{P}=\{\mathbf{p}_{r}\}_{\mathbf{r}\in\mathcal{R}}$, where $\hat{\mathbf{D}}_\theta(\mathbf{r})$ represents the expected depth. To approximate this expected depth, we integrate the sampled particles along the ray direction $\mathbf{d}_{\mathbf{r}}$.

\begin{equation}
\hat{\mathbf{D}}_\theta(\mathbf{r})=\int_{u_n}^{u_f} Q(u) \sigma_\theta(\mathbf{r}(u)) u d u
\label{eq:label5}.
\end{equation}
To address the challenges posed by large-scale driving scenes, our approach involves partitioning the scene into multiple blocks and combining them into explicit 3D geometric spaces $\mathbf{P}$ based on density point information. This differs from the methods~\cite{tancik2022block} or~\cite{meuleman2023progressively}, where multiple neural radiance fields are constructed for training and rendering. Our explicit representation allows for dynamic storage and loading of scenes, enabling real-time rendering of large-scale driving scenes.
Given that $\mathbf{P}_{i}$ represents the scene of the $i_{th}$ block, we fuse multiple blocks using $\mathbf{Merge}$ operations.

\begin{equation}
\mathbf{P}={{\mathbf{Merge}}_{\mathbf{i} \in \{1,2...N\}}(\mathbf{P}_{i} , \Phi  (\mathbf{P}_{i-1},\mathbf{P}_{i}))},
\end{equation}
where the density space $\mathbf{P}_{i}$ acts as a buffer for the nearby area and incorporates the predicted density points from the distant area with the previous scene $\mathbf{P}_{i-1}$ using function $\Phi$. 
This integration smooths block transitions and supplements density points from additional directions. Interestingly, our fusion strategy eliminates the need for overlapping regions, which reduces the number of blocks required for training compared to previous methods.

 \begin{figure}[t]
\centering
  \includegraphics[width=\linewidth]{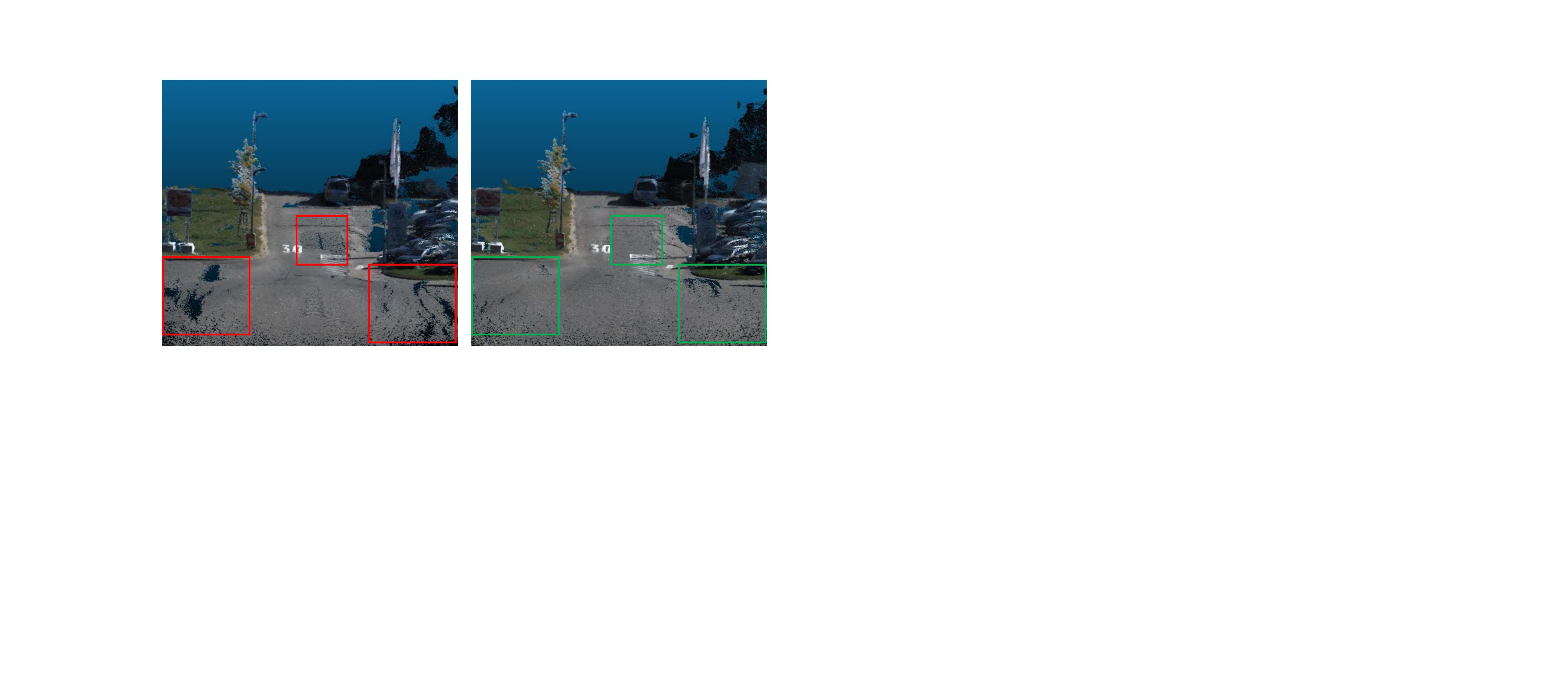}  
  \caption{Visualization of the explicit density spaces shown in point clouds to demonstrate the effectiveness of our proposed geometric regularization. The left figure shows the original result, while the right figure shows the result after applying the proposed geometric regularization.}
  \label{fig:label3}
\end{figure} 

\subsubsection{Geometric Regularization}

In driving scenes, the problem of density holes caused by sparsely observed viewpoints becomes more prominent. In order to optimize the density space, it can be seen from Eqn.~\ref{eq:label6} that the expected depth plays an important role. Therefore, we need to impose additional constraints on depth patches. Drawing inspiration from RegNeRF~\cite{niemeyer2022regnerf}, we observe that real-world geometry often exhibits piece-wise smoothing characteristics. To alleviate this, we propose a regularization method to smooth the depth of unobserved viewpoints. 
Considering the expected depth $\hat{\mathbf{D}}_\theta(\mathbf{r})$ calculated by Eqn.~\ref{eq:label4}, we formulate the depth smoothness loss as follows.
\begin{equation}
\begin{aligned}
\mathcal{L}_{\mathrm{D}}(\theta, \mathcal{R}_r) = 
&\sum_{\mathbf{r} \in \mathcal{R}_r} \sum_{i, j=1}^{S_{\mathrm {patch}}-1} \left(\hat{D}_\theta(\mathbf{r}_{ij})-\hat{D}_\theta(\mathbf{r}_{i+1j})\right)^2 \\
&+ \left(\hat{D}_\theta(\mathbf{r}_{ij})-\hat{D}_\theta(\mathbf{r}_{ij+1})\right)^2 \\
&+ \left(\hat{D}_\theta(\mathbf{r}_{ij})-\hat{D}_\theta(\mathbf{r}_{i+1j+1})\right)^2,
\end{aligned}
\label{eq:label8}
\end{equation}
where $\mathcal{R}_{r}$ represents a collection of rays sampled from camera poses. Each $\mathbf{r}_{i j}$ denotes the ray passing through pixel $(i,j)$ within a patch centered at $r$. The variable $S_{\mathrm{patch}}$ refers to the size of the rendered patches. The objective of the smoothness loss is to minimize depth discrepancies between neighboring pixels in a specific image sub-block. Fig.~\ref{fig:label3} illustrates a visualization of the density space after incorporating the depth smoothness constraint, which indicates that such constraint effectively smooths out certain unobserved regions and fills the holes.

By combining Eqn.~\ref{eq:label4} and Eqn.~\ref{eq:label8}, we obtain the following training losses to optimize the density space:
\begin{equation}
L\left(\theta,\mathcal{R}_r\right)=\mathcal{L}_{\mathrm{MSE}}\left(\theta,\mathcal{R}_r\right) + \lambda \mathcal{L}_{\mathrm{D}}\left(\theta, \mathcal{R}_r\right).
\label{eq:label9}
\end{equation}
To enhance the reconstruction quality and ensure the accuracy of the density space, we utilize the sampling technique from~\cite{tancik2023nerfstudio}. This sampling method focuses on selecting positions within the scene that heavily contribute to the final rendering. 
Instead of using position encoding in NeRF, we employ spherical harmonics (SH) functions~\cite{chen2022tensorf}. This enables encoding the ray direction via a fixed SH function, eliminating the need for a neural network representation.

\subsection{Density-Guided Differentiable Rendering}

By taking advantage of the density-guided scene representation, a 3D density space is obtained to represent the scene. This learnable density space allows for dynamic optimization, as it can be refined based on feedback from rendered images. We employ a neural renderer to synthesize photorealistic images using the neural density feature derived from it. The process involves two key modules, including Point Rasterization and Scene Optimization. These modules will be explained in detail in the following.

\subsubsection{Point Rasterization}
We employ a rasterization process where density points from the density space $\mathbf{P}$ are projected using a specified camera model $C$. The rasterization phase involves creating images of dimensions $W \times H$ based on the pinhole camera $C$. Following the approach outlined in~\cite{li2023read}, a pyramid of rasterized raw images is constructed. This assigns points that successfully pass the depth test to the neural density feature, which is subsequently projected onto the corresponding pixel using the complete projection transformation of the camera.
Given the intrinsic and extrinsic matrices of camera $C$, the first step is to project each world point in density space $\mathbf{P}$ onto the image space of layer $l$. Considering the camera model $C$ and the rigid transformation from world to camera-space $(R, T)$, we define this projection as follows.
\begin{equation}
Q_s(p_r)=\frac{1}{2 l} C(R( p_r)+T).
\end{equation}

The real-valued $p_r$ is then converted into pixel coordinates by rounding it to the nearest integer. A world point $p_r \in \mathbb{R}^3$ is therefore projected into pixel coordinates $q_r \in \mathbb{Z}^2$ by
$
q_r=\left\lfloor Q_s\left(p_r \right)\right\rceil.
$
We further encode pixels $q_r$ into neural density features $\tau$ through $1\times1$ convolution kernel. 
The neural density features $\tau$ encode the local 3D scene context around $p_{r}$. We employ a neural renderer with learnable parameter $\theta$ to project all the neural density features $\tau$ onto the RGB image space expressed as $\hat{I}$:

\begin{equation}
\hat{I} = \psi(\mathbf{P}, C, \tau, \theta),
\end{equation}
where $\psi$ is a neural rendering network (neural renderer) based on U-Net~\cite{ronneberger2015u}. Like~\cite{li2023read}, we employ multi-scale neural density features as inputs for the rendering network. By fusing features from different scales, we mitigate the problem of missing neural density features that may occur in sparse-density spaces. To further enhance the performance, we incorporate Gate convolution~\cite{yu2019free}, which effectively filters out invalid features within the neural density feature representation.

 \begin{figure}[t]
\centering
  \includegraphics[width=\linewidth]{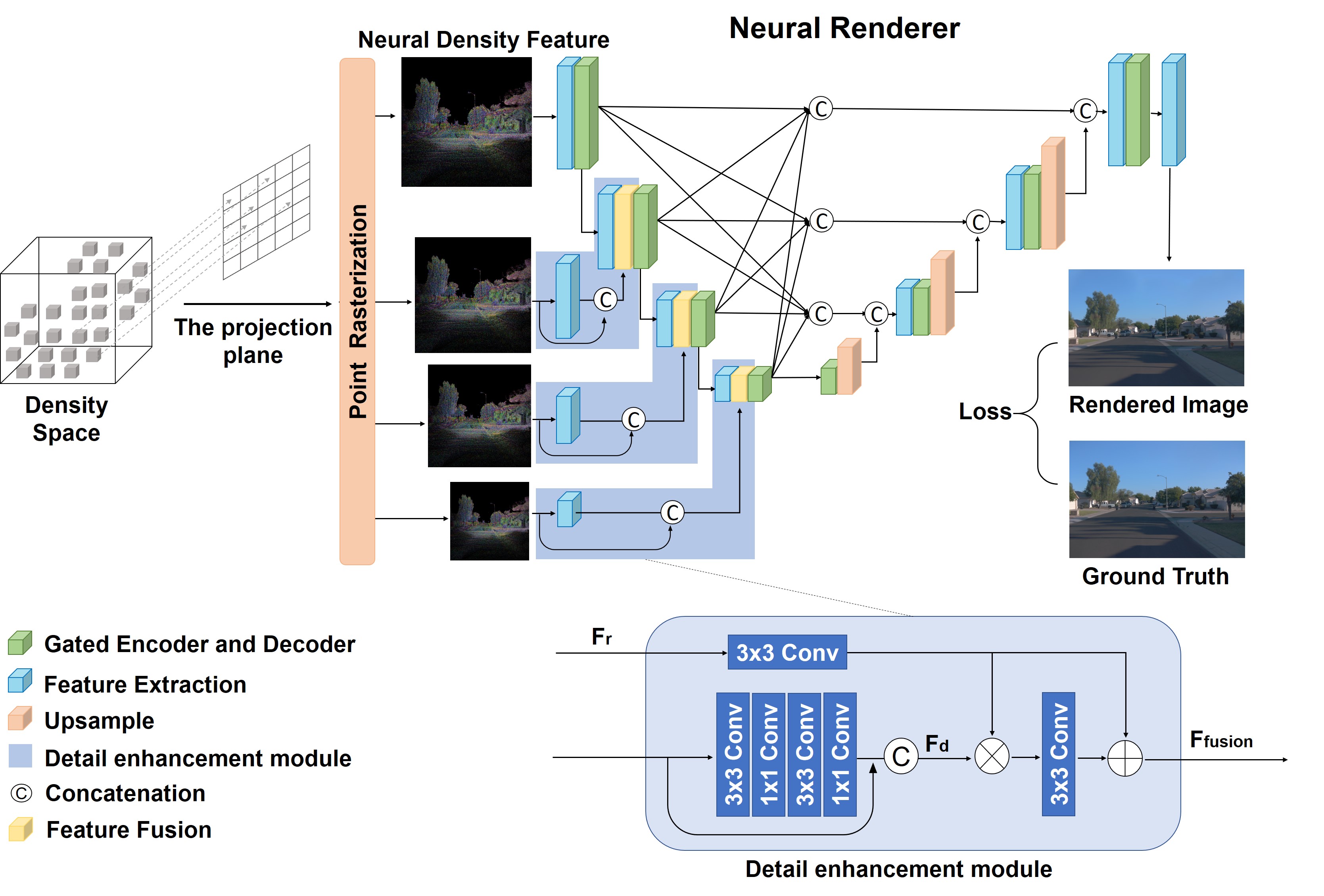}  
  \caption{Overview of our proposed Density-Guided Neural Renderer. We begin by rasterizing density points from the density space at multiple
resolutions. Subsequently, the neural renderer combines the multi-scale density features to synthesize an image in novel view..
}
  \label{fig:label8}
\end{figure}

\subsubsection{Density-Guided Neural Renderer}

In regions of the density space where there are large holes, such as the sky, vehicles, or distant areas, the density points are sparser, resulting in insufficient coverage of those areas in the image, and blurriness is usually observed.
To address this issue, we propose a neural renderer based on a multi-scale rendering structure U-Net~\cite{ronneberger2015u} as in READ~\cite{li2023read}. By trade-offing between rendering quality and efficiency, we employ a detail enhancement module composed of $3 \times 3$ and $1 \times 1$ convolution layers $CONV$ to extract the multi-scale neural density features, denoted as $F_d$. As depicted in Fig.\ref{fig:label8}, the proposed detail enhancement module takes advantage of the complementary information of fusion feature $F_{fusion}=F_r+F_r\odot F_d$, where $\odot$ denotes the element-wise multiplication. $F_r$ is the reduced feature. The neural density feature generated from the sparse density space can effectively fill the holes in the urban scene, and optimize the details of scenes.

Through the utilization of resizing, the output of the Gate encoder establishes effective interconnections among intra-scale features. This facilitates the seamless exchange of information across varying scales of features. Consequently, the multi-scale features are merged with their up-sampled counterparts, effectively addressing the gaps in the neural density features derived from the density space. This fusion process enhances the scene's level of detail and successfully improves previously blurred areas.

\subsubsection{Gate Encoder and Decoder}

To distinguish the pixels inside and outside the hole, we introduce the gated convolutions~\cite{yu2019free} into the Gate encoder and decoder, as shown in Fig.\ref{fig:label8}. This enables the Gate encoder and decoder to dynamically select features for each channel and spatial location. The selection process is defined as follows:
\begin{equation}
F_{\text {output }}=F_{\text {input }}+\phi\left(W_{f} \cdot F_{\text {input }}\right) \odot \sigma\left(W_{g} \cdot F_{\text {input }}\right)
\end{equation}
where the input feature $F_{input}$ is firstly passed through the two different convolution filters $W_{g}$ and $W_{F}$. We intend to filter the features with a mask and output threshold between 0 and 1, where the sigmoid function $\sigma$ is used. To improve the learning efficiency, we use the ELU activation function $\phi$ for the feature. $\odot$ denotes the element-wise multiplication. The filtered feature is fused with the input as a new feature.

\subsubsection{Scene Optimization}

Since the density space lacks explicit geometric supervision, it is susceptible to containing noise and missing parts of the surface. These issues may degrade the quality of rendering and potentially introduce artifacts. To address this problem, we define patches in the rendered image as ($S_{p}$) and the real image as ($S_{g}$), respectively. To measure the high-level visual similarity between these images, we make use of the perceptual loss $VGG(S_{p}, S_{g})$ that serves to estimate potential geometric quality problems within the density space. The estimation is performed based on the value of $VGG(S_{p}, S_{g})$, allowing us to identify and address problematic geometric aspects.

During the training process, we focus on patches where the value of $VGG(S_{p},S_{g})$ is lower than half the average value across the training set. This selection criterion enables us to prioritize patches with improved visual similarity. For these selected patches, we augment the density space by adding a group of points along the pixel ray. To optimize the corresponding 3D density points, we consider the depth range of neighboring pixels.
To address noise in the density space, we implement a method that assesses whether a neural density point lies near the scene's surface. This step allows us to identify and remove unnecessary outliers. We achieve this by examining the number of points within a specified radius around each neural density point. If the count falls below a predefined threshold, the point is classified as an outlier and subsequently removed from the density space. This process improves the quality of the density space and reduces the impact of noise.

\subsubsection{Loss Function}

When training novel view synthesis networks, traditional reconstruction loss measures like Mean Squared Error (MSE) or $L_1$ loss are commonly used. However, these pixel-level comparisons may not fully capture high-level visual similarity. To address this limitation, we incorporate the perceptual loss~\cite{johnson2016perceptual} into the training process. By doing so, we aim to preserve high-frequency details while promoting color fidelity in the generated images. Specifically, we compute the perceptual loss between the synthetic novel view and ground truth image $I_{GT}$, which is calculated by a pre-trained VGG layer as follows:
\begin{equation}
L\left(\tau, \theta\right)=L_{VGG}(I_{GT},\psi\left(P, C, \tau, \theta) \right).
\end{equation}
Given density space $P$ and camera parameters $C$, our neural renderer learns the neural density feature $\tau$ and network parameters $\theta$.


 \begin{figure*}[t]
\centering
  \includegraphics[width=\linewidth]{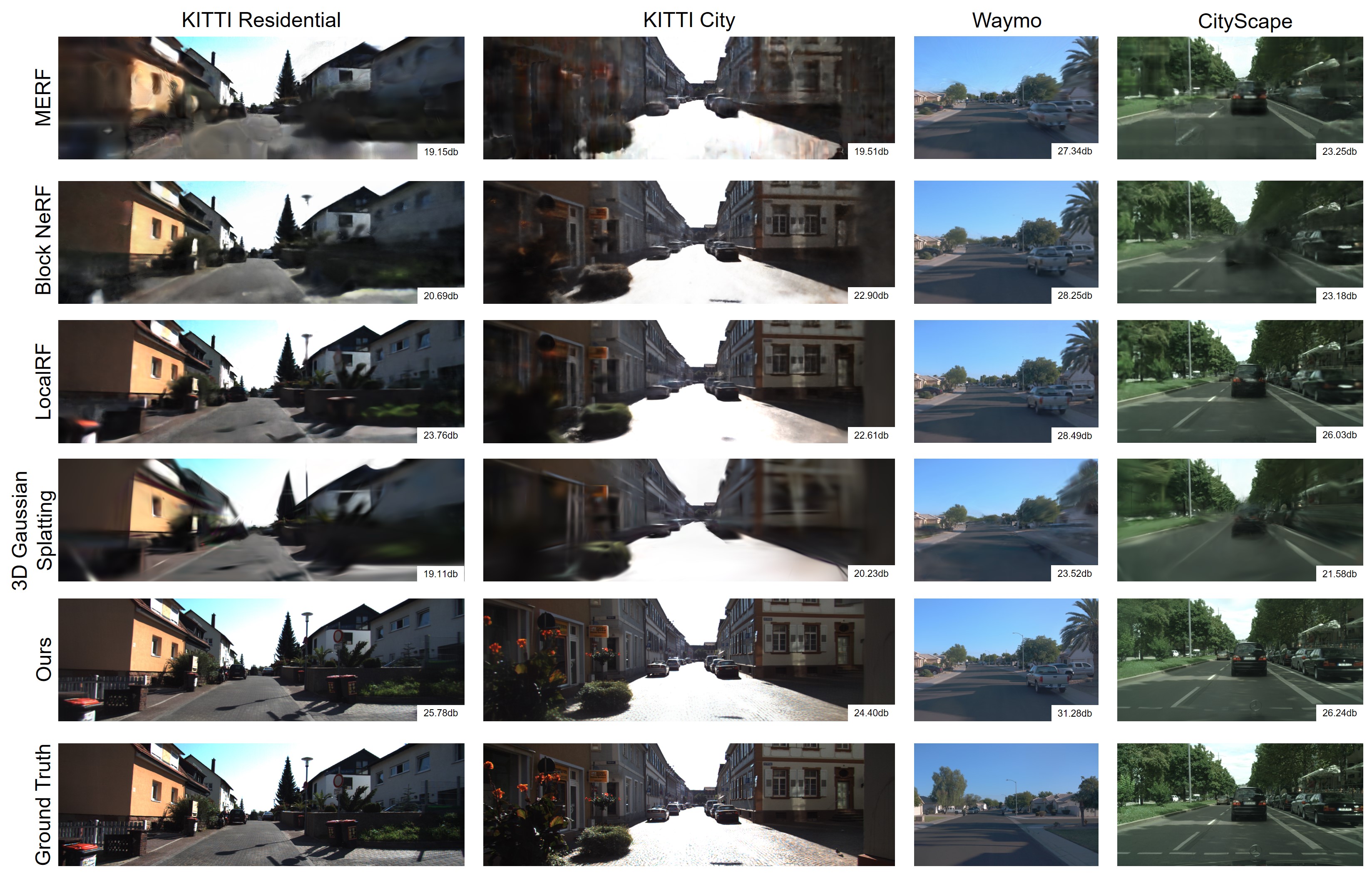}  
  \caption{Comparative results of novel view synthesis were obtained for residential and city scenes from benchmark datasets like KITTI, CityScape, and Waymo. DGNR was evaluated against existing approaches including MERF, Block-NeRF, LocalRF and 3D Gaussian Splatting. Our proposed DGNR approach outperforms the other methods in synthesizing realistic views of buildings, vehicles, and distant regions. For detailed results, please refer to the supplementary materials. Best viewed in color and zoomed in.
}
  \label{fig:label4}
\end{figure*}

\section{Experimental Results}

\subsection{Experimental Setup}

To comprehensively evaluate DGNR and assess its computational efficiency, we conducted experiments on an NVIDIA Tesla V100 16GB GPU. DGNR was rigorously tested on five datasets of driving scenes. Through a combination of qualitative and quantitative analysis, we present compelling results demonstrating the promising performance and efficiency of our method compared to other approaches.
During training, we employed the Adam optimizer with a learning rate of $0.001$ to optimize the network parameters. We adopted a decay strategy that gradually reduced the learning rate from its initial value to $10^{-6}$ throughout training. Additionally, the remaining parameters are as follows. The depth patch size $S_{\mathrm{patch}}$ was set to $8\times8$, and $\lambda$ was chosen to be $10^{-5}$ to balance the impact of different losses and regularization terms.

\subsection{Testbeds}

In our experiments, we selected datasets originally designed to capture the complexities of large-scale street scenes with long trajectories. To this end, we made use of widely recognized datasets like KITTI~\cite{geiger2012we} and Cityscapes~\cite{cordts2016cityscapes}. These choices were deliberate as many existing street scene datasets have inherent limitations that hinder their suitability for synthesizing novel views along extended trajectories. For example, certain datasets prioritize visual diversity over maintaining coherent street scene sequences, as in the case of NuScenes~\cite{caesar2020nuscenes}. Conversely, other datasets focus on shorter trajectories rather than long ones, such as the Waymo Open Dataset~\cite{sun2020scalability} and Argoverse~\cite{chang2019argoverse}. We carefully considered these limitations and opted for datasets that aligned well with our objective of synthesizing novel views in large-scale driving scenes with long trajectories.

\textbf{KITTI}~\cite{geiger2012we}: Our experiments were conducted on three different scenarios, all extracted from the KITTI dataset. These scenarios comprised a total of 921 frames for the Residential scene, 819 frames for the Road scene, and 1584 frames for the City scene. We follow the training and testing split in~\cite{meuleman2023progressively,tancik2022block}, testing every ten frames (e.g., frame 0, 10, 20...) while utilizing the remaining frames for training.

\textbf{Cityscapes}~\cite{cordts2016cityscapes}: To evaluate more complex street scenes, we chose a large-scale dataset featuring diverse sequences captured across 50 different cities. This dataset provides a more complex environment and showcases images with a resolution of 2048x1024. From this dataset, we selected a specific video sequence comprising 1100 frames. To maintain consistency and facilitate comparisons with the KITTI dataset, we followed the same testing frame selection as the KITTI dataset. This ensures that the test frames align across both datasets, enabling a fair evaluation of our method's performance in various street scenes.

\begin{table*}[th]
\centering
\caption{Quantitative evaluation of novel view synthesis on the KITTI dataset. PCI: Point Cloud Initialization.}
\begin{tabular}{lcccc}
\toprule
Method & PCI & KITTI Residential & KITTI Road & KITTI City\\
& & VGG$\downarrow$ / PSNR$\uparrow$ / LPIPS$\downarrow$ / SSIM$\uparrow$ & VGG$\downarrow$ / PSNR$\uparrow$ / LPIPS$\downarrow$ / SSIM$\uparrow$ & VGG$\downarrow$ / PSNR$\uparrow$ / LPIPS$\downarrow$ / SSIM$\uparrow$\\
\midrule
NGP~\cite{muller2022instant} 
& \XSolidBrush 
& 1068.9 / 15.12 / 0.7836 / 0.4769 
& 989.7 \enspace / 17.85 / 0.6736 / 0.5021 
& 1186.0 / 14.12 / 0.8263 / 0.4318 \\
NGP+Cont 
& \XSolidBrush 
& 1006.2 / 16.81 / 0.6070 / 0.5258 
& 934.7 \enspace / 18.06 / 0.5629 / 0.5303 
& 1162.7 / 15.19 / 0.6816 / 0.4621 \\
MERF~\cite{reiser2023merf} 
& \XSolidBrush 
& 990.9 \enspace / 17.52 / 0.5610 / 0.5373 
& 901.5 \enspace / 19.54 / 0.5374 / 0.5425 
& 1134.6 / 15.62 / 0.6191 / 0.4797 \\
Block-NeRF~\cite{tancik2022block} 
& \XSolidBrush 
& 818.4 \enspace / 20.22 / 0.4898 / 0.6160 
& 733.7 \enspace / 21.43 / 0.4404 / 0.6266 
& 818.8 \enspace / 20.55 / 0.4411 / 0.6465 \\
LocalRF~\cite{meuleman2023progressively} 
& \XSolidBrush 
& 788.9 \enspace  / 20.72 / 0.4662 / 0.6392 
& 745.7 \enspace / 20.97 / 0.4512 / 0.6287 
& 831.1 \enspace / 20.14 / 0.4373 / 0.6445 \\
3D Gaussian Splatting~\cite{kerbl20233d} 
& \XSolidBrush 
& 1027.9 / 16.48 / 0.6808 / 0.5387 
& 1008.4 / 15.32 / 0.6845 / 0.4842 
& 1118.3 / 15.24 / 0.7174 / 0.5005 \\
\textbf{Ours} 
& \XSolidBrush 
& \textbf{472.6 } / \textbf{22.00} / \textbf{0.1742} / 0.6964 
& \textbf{437.7 } / \textbf{22.94} / \textbf{0.1679} / 0.6857 
& \textbf{434.8 } / \textbf{22.80} / \textbf{0.1401} / \textbf{0.7529} \\
\midrule
READ~\cite{li2023read} 
& \Checkmark 
& 522.0 \enspace / 21.55 / 0.2164 / \textbf{0.7076} 
& 510.9 \enspace / 22.05 / 0.2019 / \textbf{0.6965} 
& 563.6 \enspace / 21.14 /  0.2121 /  0.7062\\
3D Gaussian Splatting~\cite{kerbl20233d} 
& \Checkmark 
& 890.7 \enspace / 19.09 / 0.5748 / 0.6081 
& 789.0 \enspace / 20.47 / 0.4663 / 0.6174 
& 1003.2 / 18.26 / 0.6154 / 0.5723 \\
\bottomrule
\end{tabular}
\label{tab:table1}
\end{table*}

\begin{table*}[th]
\centering
\caption{Quantitative evaluation of novel view synthesis on the CityScape and Waymo dataset. Note that Cityscape dataset does not provide LiDAR point cloud to initialize READ and 3D Gaussian Splatting.}
\begin{tabular}{lccc}
\toprule
Method & PCI & CITYSCAPE & WAYMO \\
& & VGG$\downarrow$ / PSNR$\uparrow$ / LPIPS$\downarrow$ / SSIM$\uparrow$ & VGG$\downarrow$ / PSNR$\uparrow$ / LPIPS$\downarrow$ / SSIM$\uparrow$ \\
\midrule
NGP~\cite{muller2022instant} & \XSolidBrush & 662.1 / 20.22 / 0.6772 / 0.6866 & 331.8 / 31.75 / 0.2902 / 0.8892 \\
NGP+Cont & \XSolidBrush & 616.0 / 23.30 / 0.5943 / 0.7233 & 330.5 / 31.36 / 0.2898 / 0.8909 \\
MERF~\cite{reiser2023merf} & \XSolidBrush & 610.4 / 23.52 / 0.5283 / 0.7226 & 317.3 / 31.31 / 0.2345 / 0.8880 \\
Block-NeRF~\cite{tancik2022block} & \XSolidBrush & 576.4 / 23.99 / 0.5471 / 0.7360 & 367.4 / 29.75 / 0.3397 / 0.8646 \\
LocalRF~\cite{meuleman2023progressively} & \XSolidBrush & 504.2 / 26.26 / 0.4499 / 0.7850 & 339.8 / 30.71 / 0.2997 / 0.8838 \\
3D Gaussian Splatting~\cite{kerbl20233d} & \XSolidBrush & 619.5 / 20.08 /	0.5700 / 0.7078  & 415.4 / 25.46 / 0.3648 / 0.8525  \\
\textbf{Ours} & \XSolidBrush & \textbf{329.9} / \textbf{26.64} / \textbf{0.2050} / \textbf{0.8136} & \textbf{177.5} / \textbf{32.63} / \textbf{0.1033} / \textbf{0.9087} \\
\midrule
READ~\cite{li2023read} & \Checkmark & N/A & 262.1 / 25.25 / 0.1834 / 0.8733 \\
3D Gaussian Splatting~\cite{kerbl20233d} & \Checkmark & N/A & 295.4 / 31.71 / 0.2248 / 0.9107\\
\bottomrule
\end{tabular}
\label{tab:table2}
\end{table*}

\subsection{Evaluation Results}
To demonstrate the efficacy of presented DGNR method, we compare it with the neural radiance approach for unbounded scenes without geometric prior or supervision, including Instant NGP~\cite{muller2022instant}, NGP+Contraction, MERF~\cite{reiser2023merf}, Block-NeRF (an approach that employs block-wise processing~\cite{tancik2022block} based on Instant NGP~\cite{muller2022instant}) and LocalRF~\cite{meuleman2023progressively}. These approaches have demonstrated promising results in scene synthesis of large-scale driving scenes.
To showcase the reliability of the presented DGNR method and highlight the benefits of the density space generated by it. We also compare with methods that require point clouds as initialization, namely 3D Gaussian Splatting~\cite{kerbl20233d} and READ~\cite{li2023read}, where the point clouds are scanned by LiDAR.

Similar to the evaluation protocols used in those methods, we adopt Peak Signal-to-Noise Ratio (PSNR), Structural Similarity (SSIM), Learned Perceptual Image Patch Similarity (LPIPS), and Perceptual loss (VGG loss) as the evaluation metrics. Furthermore, we offer a detailed explanation of the experimental setup for the comparative methods, ensuring transparency and clarity.

\textbf{Instant NGP}~\cite{muller2022instant} reconstructs a volume radiance field using hash encoding. We adopt the same camera parameters in our experiment to ensure a fair comparison. However, we found that Instant NGP faces challenges in synthesizing novel views of large scenes. We combine Instant NGP with a spatial-warping method, similar to the one used in MERF~\cite{reiser2023merf}, and refer to it as NGP+Cont. 

\textbf{MERF}~\cite{reiser2023merf} offers a memory-efficient representation of radiance fields for real-time rendering in large-scale scenes. This method reduces memory consumption using a sparse feature grid and high-resolution 2D feature planes. Additionally, MERF presents a contraction function to support large-scale unbounded scenes.

\textbf{LocalRF}~\cite{meuleman2023progressively} models the large-scale unbounded scenes through a joint pose and radiance field estimation method, which enables the progressive processing of video sequences by utilizing overlapping local radiance fields. To effectively capture the complexity of large-scale unbounded scenes, LocalRF dynamically instantiates local radiance fields, adapting to the scene's varying characteristics and ensuring accurate representation throughout the reconstruction process.

Since we only tested the static driving scenes in our experience, we utilized the monocular depth supervision from LocalRF and did not employ optical flow supervision to distinguish dynamic objects. We used the same camera parameters to maintain experimental fairness and did not perform pose optimization, as described in LocalRF. This is because the pose optimization in~\cite{meuleman2023progressively} is unsuitable for driving scenes captured in driving mode, where the scene sparsity poses challenges for accurate pose estimation.

\textbf{Block-NeRF}~\cite{tancik2022block} is similar to LocalRF, which effectively synthesizes large-scale driving scenes by dividing them into blocks. Due to the authors' implementation of Block-NeRF being publicly unavailable, we built a baseline using the PyTorch version of Instant NGP. To further enhance the results, we employed K-means clustering to partition the whole scene into smaller blocks. Subsequently, we trained each local region of the scene in parallel. This implementation improved the original results and leveraged the hashing encoding of Instant NGP to improve the training speed.

\textbf{3D Gaussian Splatting}~\cite{kerbl20233d} introduces an innovative approach using anisotropic 3D Gaussians as a high-quality representation of radiance fields. Since this method can randomly initialize point clouds and adaptively optimize the density of Gaussians, we conducted experiments with and without point cloud initialization.

\textbf{READ}~\cite{li2023read} presents a large-scale neural rendering method for synthesizing autonomous driving scenes, which performs well in driving scenarios. It not only enables the synthesis of realistic driving scenes but also facilitates their stitching and editing. We used the same number of U-net layers for fair comparison.

These methods can be classified as using spatial warping to handle large scenes (NGP, NGP+Cont, MERF), requiring geometric priors or point cloud initialization(READ, 3D Gaussian Splatting), and relying on scene partitioning (Block-NeRF, LocalRF, and our method). The results presented in Table~\ref{tab:table1} and Table~\ref{tab:table2} reveal that the spatial-warping-based method encounters difficulties when dealing with long urban scene trajectories. It performs comparatively worse than the methods employing scene division when evaluated across various metrics.

READ~\cite{li2023read} and 3D Gaussian Splatting~\cite{kerbl20233d} are methods that require point cloud initialization. While READ performs well in the KITTI dataset, it faces challenges in synthesizing novel views for distant, elevated, and sky regions due to the limitations of the point clouds obtained from LiDAR scans. In contrast, our density space overcomes this limitation by learning point cloud representations of these regions through scene optimization from images. This enables us to achieve improved results in synthesizing novel views of these challenging areas. It is worth mentioning that although the 3D Gaussian Splatting method can adaptively optimize the Gaussians, it may not be effective in outdoor areas, especially in sparse driving scenes. In addition, since there is no lidar data in the CITYSCAPE dataset, the 3D Gaussian Splatting and READ methods cannot evaluate the performance on this dataset.

\begin{figure*}
\centering
  \includegraphics[width=\linewidth]{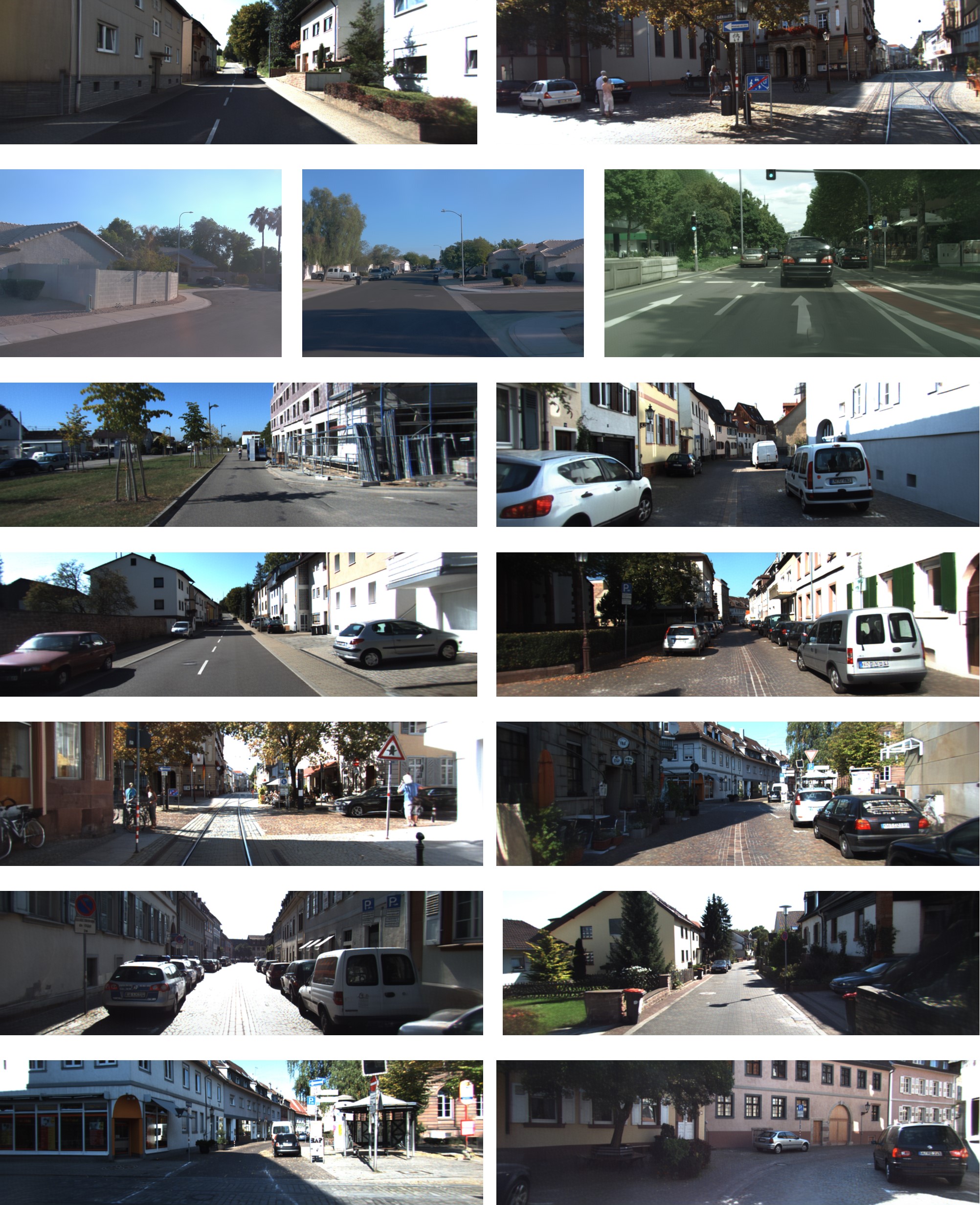}  
  \caption{More synthesized driving scenes using our proposed method.
}
  \label{fig:label7}
\end{figure*}

Block-NeRF~\cite{tancik2022block} and LocalRF~\cite{meuleman2023progressively} effectively address the synthesis of large-scale photorealistic scenes by dividing driving scenes into blocks. However, these methods tend to produce blurred results with noticeable artifacts, particularly in close-up and surrounding environments, as illustrated in Fig.~\ref{fig:label4}. Additional examples of driving scenes synthesized using our approach are shown in Fig.~\ref{fig:label7}. Our proposed method significantly outperforms Block-NeRF and LocalRF across all evaluation metrics, as shown in Table~\ref{tab:table1} and Table~\ref{tab:table2}.

To demonstrate the reliability of our approach, we conducted experiments on a scenario using the commonly-used Waymo dataset, as outlined in Table~\ref{tab:table3}. Despite scenes with fewer than 200 frames, our metrics consistently surpassed other methods, achieving a 127.0\% reduction in LPIPS error and a 78.8\% reduction in VGG error compared to without geometric prior or supervision methods.

\begin{table}[th]
\centering
\caption{Ablation study of our method on KITTI Road dataset. `B' denotes using block representation; `R' denotes regularization; `C' denotes using the completion.}
\begin{tabular}{ccc|cccc}
\toprule
B & R & C & VGG$\downarrow$ &   PSNR$\uparrow$ & LPIPS $\downarrow$ & SSIM$\uparrow$ \\
\midrule
 & & & 640.4  & 19.40 & 0.2835 & 0.5793   \\
\Checkmark & &    &  486.2  & 22.09  & 0.1907 & 0.6596  \\
\Checkmark &\Checkmark &   & 453.0  & 22.66  & 0.1767  & 0.6789 \\
\Checkmark &\Checkmark &\Checkmark  & \textbf{437.7} &   \textbf{22.94}  & \textbf{0.1679} &  \textbf{0.6857} \\             
\bottomrule
\end{tabular}

\label{tab:table3}
\end{table}

\begin{table}[!t]
\centering
\caption{Comparisons of the rendering speed on KITTI dataset.}
\begin{tabular}{lcccc}
\toprule
Method  &   PSNR $\uparrow$ &  LPIPS $\downarrow$ & Model Size $\downarrow$ & FPS$\uparrow$  \\
\midrule
Block-NeRF~\cite{tancik2022block} &  21.43 &  0.4404 &  4121MB & 0.82  \\
LocalRF~\cite{meuleman2023progressively} & 20.97 & 0.4512 & 4331MB & 0.13 \\
\textbf{Ours} &  \textbf{22.94}  &  \textbf{0.1679}  & \textbf{237MB}  & \textbf{16.67}  \\
\bottomrule
\end{tabular}
\label{tab:table4}
\end{table}

\subsection{Ablation Study}

In this part, we conduct qualitative and quantitative experiments to evaluate each component of our method on KITTI Road dataset.
As a baseline, we initially train the entire scene without any geometric optimization, shown in the first row of Table~\ref{tab:table3}. To enhance the geometry quality, we introduce Density-Based Scene Fusion to further optimize the representation. This involves using smaller blocks and a learnable density-guided fusion, significantly enhancing the scene's geometric quality and improving the fidelity of the rendered images.
Additionally, we incorporate geometric regularization to optimize the scene geometry. The quantitative results demonstrate a slight performance improvement. Moreover, we employ density-guided differentiable rendering techniques to complete the density points by feeding the predicted rendering images back to the density space. This module effectively mitigates the blurriness issue encountered in novel view synthesis, as depicted in Fig.~\ref{fig:label4}.

 \begin{figure*}[t]
\centering
  \includegraphics[width=\linewidth]{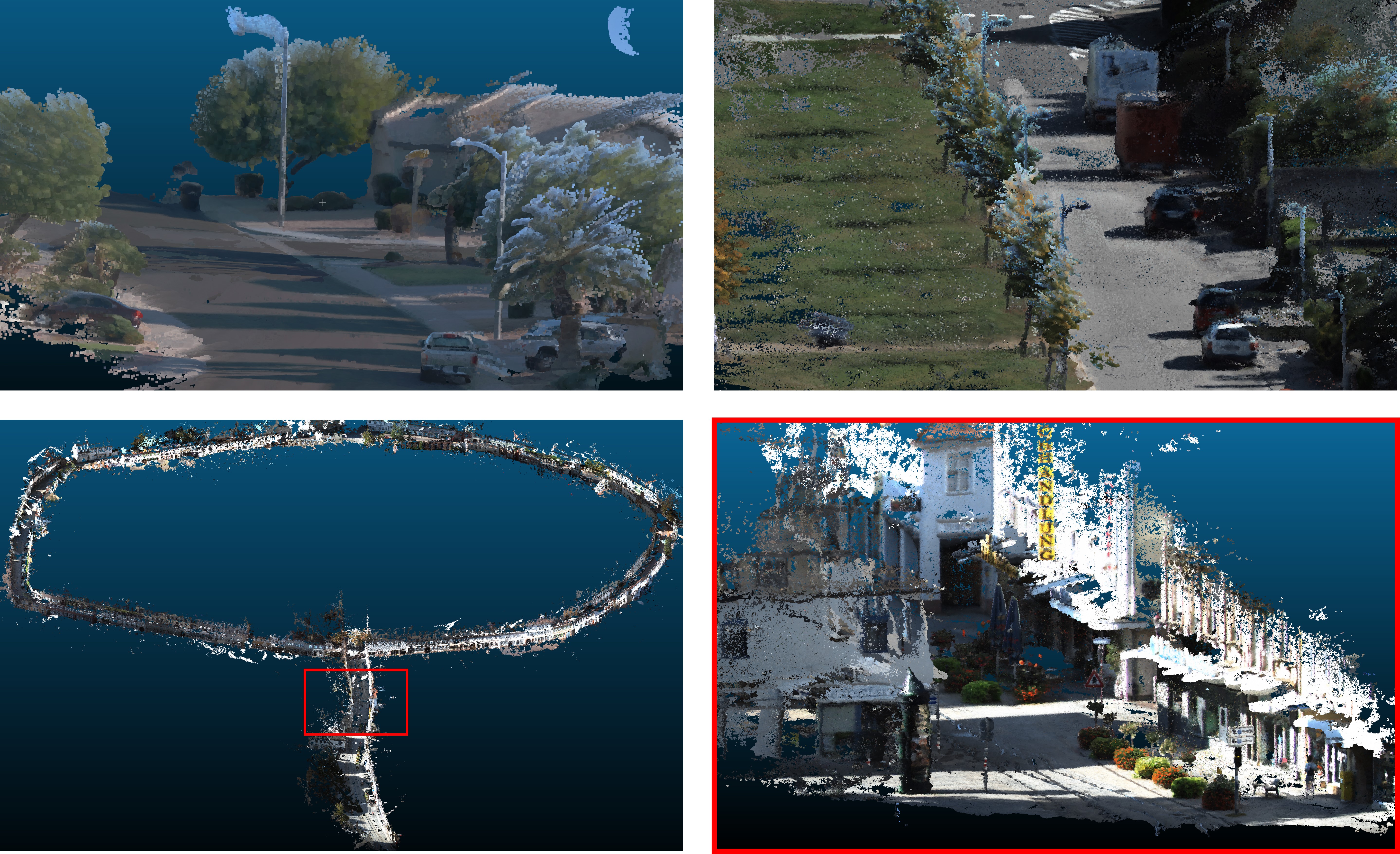}  
  \caption{Visualization of scene geometry.
}
  \label{fig:label5}
\end{figure*}

 \begin{figure}[t]
\centering
  \includegraphics[width=\linewidth]{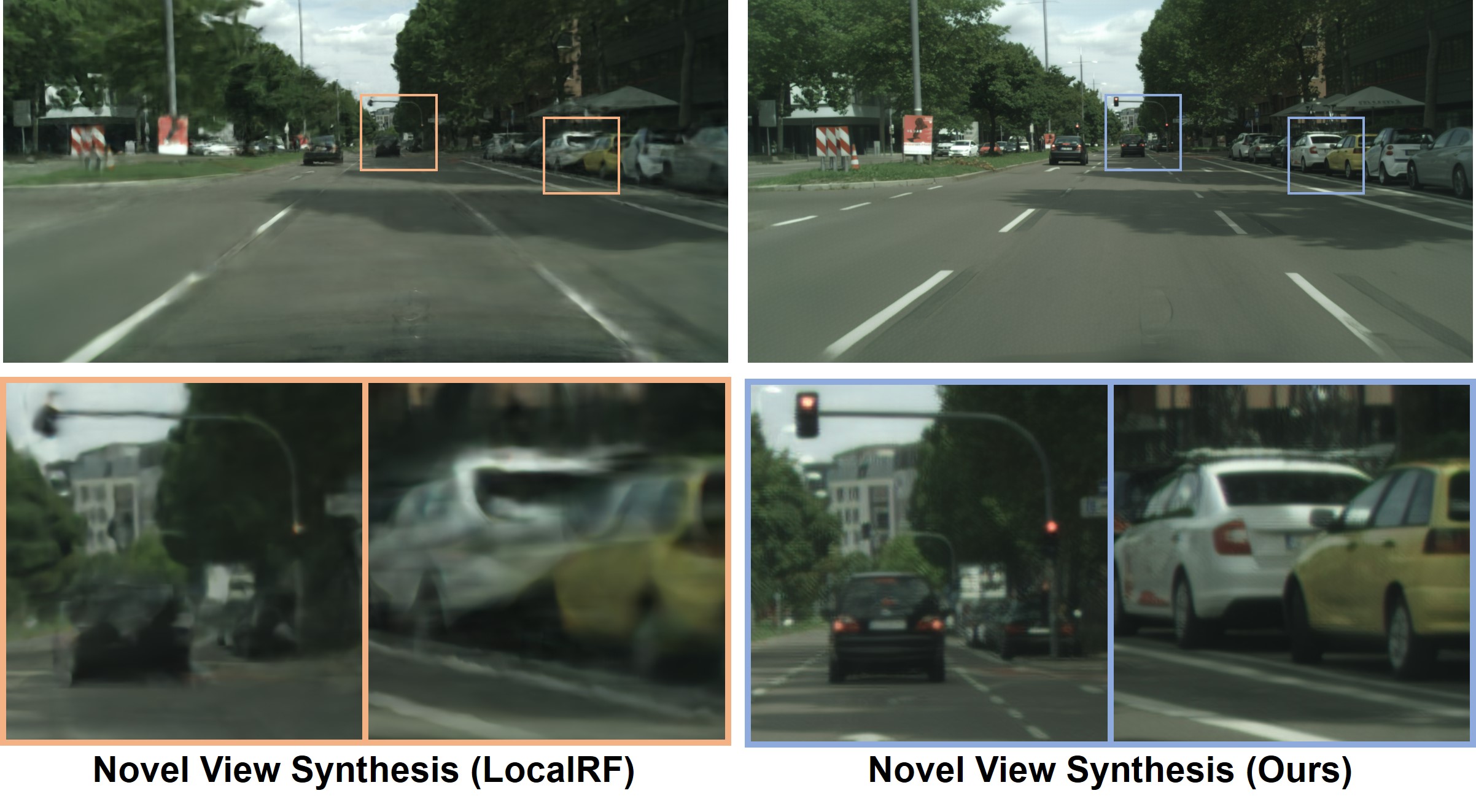}  
  \caption{Comparison with novel view synthesis using LocalRF method. Please zoom in for detailed results. 
}
  \label{fig:label6}
\end{figure}

\subsection{Complexity Analysis}

\subsubsection{Rendering Speed}

Rendering speed plays a crucial role in the synthesis of novel views, as it directly impacts interaction efficiency, which is important for certain applications such as autonomous driving simulation or virtual reality. We evaluate the rendering times of various methods without geometric priors or supervision that perform well on large driving scenes, such as Block-NeRF and LocalRF, as shown in Table~\ref{tab:table4}. Our method outperforms the others in terms of both quality and speed. Note that our implementation is a proof-of-concept without engineering optimization. 
In specific, our approach achieves rendering speeds that are $20\times$/$128\times$ than Block-NeRF and LocalRF, respectively. This advantage stems from the fact that volumetric rendering requires a large amount of computational burdens by sampling spatial points and evaluating networks. In contrast, our method eliminates the need for volume rendering during the inference stage. Instead, we leverage differentiable point rendering directly from the explicit density space. This significantly enhances rendering efficiency without compromising quality.

\subsubsection{Model Size}

Block-NeRF and LocalRF rely on radiance fields that are composed of multiple blocks. As the size of the scene increases, the model sizes of Block-NeRF and LocalRF also grow substantially to 4121MB and 4331MB, respectively. In contrast, our explicit representation approach leverages efficient differentiable rendering during the rendering stage, resulting in a significantly reduced model storage space requirement.
Specifically, our model comprises a neural renderer (U-Net) and a neural density feature that represents the density space. The sizes of these components are comparatively smaller, with the neural renderer occupying 116MB and the neural density feature occupying 121MB. By utilizing this explicit representation and compact model design, we achieve efficient storage without compromising the effectiveness of our method.

\subsubsection{Density Space Visualization}

We reconstruct an urban scene from the KITTI dataset that is nearly 1,600 frames long, covering about 1,340 meters. To demonstrate the effectiveness of our proposed geometric regularization approach, we visualized the explicit density space displayed in the point cloud. As shown in Fig.~\ref{fig:label5}, the reconstructed point cloud retains fine details and has a consistent structure over an extensive area.

\subsubsection{Discussion on Texture Quality}

To demonstrate the effectiveness of geometric optimization in our method, we show geometric details tested in Waymo and KITTI data, as illustrated in Fig.~\ref{fig:label5}. For more results please refer to the supplementary material. We discuss the texture quality from the perspective of evaluation metrics. On the Cityscape dataset (Table~\ref{tab:table2}), our method shows similar PSNR values compared to LocalRF. However, compared to LocalRF, our method reduces errors in VGG and LPIPS metrics by 119.4\% and 52.8\%, respectively. This discrepancy arises because PSNR focuses on pixel-level details through mean squared error analysis, while VGG and LPIPS metrics measure overall image similarity. In Fig.~\ref{fig:label6}, our method exhibits prominent texture details compared to other methods using volume rendering techniques.

\begin{table*}[th]
\centering
\caption{Comparison results of different scene sizes synthesized by various methods on KITTI dataset.}
\begin{tabular}{lcccc}
\toprule
   Method & PCI & Scene 1(200 frames) & Scene 2(400 frames) & Scene 3(800 frames)\\
&   & VGG$\downarrow$ / PSNR$\uparrow$ / LPIPS$\downarrow$ /  SSIM$\uparrow$
   & VGG$\downarrow$ / PSNR$\uparrow$ / LPIPS$\downarrow$ /  SSIM$\uparrow$
   & VGG$\downarrow$ / PSNR$\uparrow$ / LPIPS$\downarrow$ /  SSIM$\uparrow$  \\
\midrule
NGP~\cite{muller2022instant}  
& \XSolidBrush 
& 799.6 / 21.61 / 0.4085 / 0.6416 
& 961.3 / 19.20 / 0.5616 / 0.5665 
& 1075.6 / 16.44 / 0.7099 / 0.4959\\
NGP+Cont 
& \XSolidBrush 
& 730.6 / 20.73 / 0.3499 / 0.6664 
& 883.7 / 18.40 / 0.4619 / 0.5924 
& 1070.3 / 16.55 / 0.6167 / 0.5023 \\
MERF~\cite{reiser2023merf} 
& \XSolidBrush 
& 635.0 / 22.67 / 0.2748 / 0.7105 
& 759.6 / 21.07 / 0.3639 / 0.6647 
& 1022.0 / 16.96 / 0.5307 / 0.5323 \\
Block-NeRF~\cite{tancik2022block} 
& \XSolidBrush 
& 843.9 / 21.62 / 0.4740 / 0.6680 
& 808.3 / 21.63 / 0.4488 / 0.6691 
& 791.9 / 21.30 / 0.4365 / 0.6681 \\
LocalRF~\cite{meuleman2023progressively} 
& \XSolidBrush 
& 773.9 / 21.08 / 0.4145 / 0.6580 
& 772.5 / 21.26 / 0.4135 / 0.6629 
& 784.4 / 20.86 / 0.4214 / 0.6621 \\
3D Gaussian Splatting~\cite{kerbl20233d} 
& \XSolidBrush 
& 734.3 /	19.62 /	0.3595 /	0.6534 
& 823.7 /	19.43 /	0.4426 /	0.6375 
& 974.4 / 17.87 / 0.5970 / 0.5758 \\
\textbf{Ours} 
& \XSolidBrush 
& \textbf{358.6} / \textbf{24.30} / \textbf{0.1048} / \textbf{0.7918} 
& \textbf{370.9} / \textbf{24.04} / \textbf{0.1152} / \textbf{0.7910} 
& \textbf{383.8} / \textbf{23.67} / \textbf{0.1199} / \textbf{0.7793} \\
\midrule
READ~\cite{li2023read} 
& \Checkmark 
& 480.8 / 21.98 / 0.1673 / 0.7612
& 485.9 / 22.31 / 0.1710 / 0.7503
& 505.0 / 22.08 / 0.1806 / 0.7350 \\
3D Gaussian Splatting~\cite{kerbl20233d} 
& \Checkmark
& 600.0 / 21.45 / 0.2431 / 0.7180
& 688.3 / 21.03 / 0.3297 / 0.6955
& 856.7 / 19.73 / 0.4781 / 0.6315\\
\bottomrule
\end{tabular}
\label{tab:table5}
\end{table*}

\subsection{Influence of Scene Size}
In this section, we conduct experiments using scenes of different scales to gain a deeper understanding of how varying scene sizes impact the performance of existing methods. This analysis provides valuable insights into the challenges and limitations associated with handling longer sequences within urban environments.

To perform these experiments, we utilized the KITTI City dataset. We partitioned the dataset into three scenes, each containing a different number of frames. The first scene (Scene 1) consisted of 200 frames, the second scene (Scene 2) had 400 frames, and the third scene (Scene 3) contained 800 frames.

As shown in ``Scene 1" in Table~\ref{tab:table5}, the NGP, NGP+Cont, and MERF methods exhibit effective NeRF representations, outperforming those block-based methods like Block-NeRF and LocalRF in short sequence scenes. However, the situation reverses as the scene size increases to 400 frames. Block-based methods begin to demonstrate their advantages. With the increasing size of the scene, Block-NeRF, LocalRF, and our proposed method maintain stable performance, while NGP, NGP+Cont, and MERF methods show significant decreases in metrics. This highlights the effectiveness of scene partitioning strategies in larger scenes and the limitations of neural radiance fields in representing overly large scenes.

It is worth noting that there is no significant improvement in the performance of Block-NeRF and LocalRF methods when the scene size is reduced to 200 frames in ``Scene 1". This can be attributed to these methods relying on multiple radiance fields, which may lead to inconsistencies in smaller scenes. In contrast, our proposed method utilizes a density space representation, demonstrating consistent performance across different scene scales. This showcases the robustness of our density space in effectively representing scenes, regardless of their size, while maintaining real-time rendering.

Through these experiments, we validate the effectiveness of our proposed method specifically designed for long sequences in driving scenes.

\begin{table}[th]
\centering
\caption{Quantitative evaluation of different dividing strategies for novel view synthesis on KITTI Road dataset.}
\begin{tabular}{lcccc}
\toprule
Method  & VGG$\downarrow$ & PSNR$\uparrow$ & LPIPS$\downarrow$ & SSIM$\uparrow$ \\
\midrule
w/o block & 640.4 & 19.40 & 0.2835 & 0.5793 \\
block w/ 200 & 533.7 & 21.16 & 0.2162 & 0.6310 \\
block w/ 100 & \textbf{486.2} & \textbf{22.09} & \textbf{0.1907} & \textbf{0.6596} \\
\bottomrule
\end{tabular}
\label{tab:table6}
\end{table}

\subsection{Influence of scene division}

In this section, we delve into the process of scene division in our method and analyze the effects of employing different division strategies on the results. Notably, this experiment specifically excludes any form of geometric optimization to isolate the influence of scene division alone.

Specifically, we establish a baseline method without any scene division strategy, as shown in the first row of Table~\ref{tab:table6}. Subsequently, we divide the scene into blocks with capacities of 200 frames and 100 frames. These divisions are referred to as ``w/o block", ``block w/ 200", and ``block w/ 100", respectively. As shown in Table~\ref{tab:table6}, it can be observed that smaller scene divisions lead to more detailed geometric optimization of the scene, resulting in an improvement in the quality of the rendered image. In our main manuscript, we chose the ``block w/ 100" division strategy, as it demonstrated favorable results in terms of image quality and optimization.

\section{Conclusion}
This paper proposed an effective density-guided neural rendering approach for large-scale driving scenes. Unlike existing methods, we introduced density-guided large-scale scene representation without relying on geometric priors. Through learnable density-guided fusion and geometric regularization, our method generated more accurate geometric representations, which can be efficiently rendered through differentiable point-based rasterization. The extensive experiments on several autonomous driving datasets demonstrated the efficacy of our proposed approach in synthesizing the photorealistic driving scenes and achieving real-time capable rendering.

\newpage

\vfill

\end{document}